\documentclass[10pt, a4paper]{article}
\usepackage{lrec2016}
\usepackage{url}
\usepackage{latexsym}
\usepackage{booktabs}
\usepackage{graphicx}
\usepackage{enumitem}
\usepackage{gb4e}

\newlist{inlinelist}{enumerate*}{1}
\setlist*[inlinelist,1]{%
  label=(\roman*),
}

\title{Sieve-based Coreference Resolution in the Biomedical Domain}

\name{Dane Bell, Gus Hahn-Powell, Marco A. Valenzuela-Esc\'{a}rcega, Mihai Surdeanu}

\address{University of Arizona \\
		Tucson, AZ 85721, USA \\
		\{dane, hahnpowell, marcov, msurdeanu\}@email.arizona.edu\\}

\abstract{We describe challenges and advantages unique to coreference resolution in the biomedical domain, and a sieve-based architecture that leverages domain knowledge for both entity and event coreference resolution. Domain-general coreference resolution algorithms perform poorly on biomedical documents, because the cues they rely on such as gender are largely absent in this domain, and because they do not encode domain-specific knowledge such as the number and type of participants required in chemical reactions. Moreover, it is difficult to directly encode this knowledge into most coreference resolution algorithms because they are not rule-based. Our rule-based architecture uses sequentially applied hand-designed ``sieves'', with the output of each sieve informing and constraining subsequent sieves. This architecture provides a 3.2\% increase in throughput to our Reach event extraction system with precision parallel to that of the stricter system that relies solely on syntactic patterns for extraction.
\\ \newline \Keywords{coreference resolution, information extraction, biomedical text mining}}

\begin{document}

\maketitleabstract

\section{Introduction}


Cancer is triggered by extremely complex protein signaling pathways that interact in an intricate network, leading to accordingly complex reactions to cancer treatment drugs. This complexity, combined with the rapid pace of publishing in the biomedical domain, makes these systems effectively intractable for human experts, and so studies are typically reductionist, focusing on a single pathway at a time.

The Defense Advanced Research Projects Agency (DARPA) created the Big Mechanism program \cite{cohen2015bmp} to develop a holistic view of these cancer pathways through automated, large-scale reading and pathway assembly in amendment of and extension to already-existing human-made models of protein reaction networks. Two major requirements to accomplish this ambitious goal are effective automatic reading and assembly, i.e., extracting individual biochemical interactions, linking these isolated interactions together to form larger pathways, and merging them into an existing cancer model. 

Coreference resolution is a fundamental requirement for effective reading and assembly, as it is crucial for deciding which spans of text refer to the same entity in the real world, which can have such disparate labels as ``the protein'', ``both'', or ``ASPP1''. 
However, with the notable exception of the BioNLP shared task~\cite{kim2013ge}, coreference has rarely been a focus of reading-based work.

Domain-general coreference systems perform poorly in this domain, because they fail to capitalize on domain-specific constraints on possible coreference relations. This is illustrated in example~\ref{ex:ds-selfbindingwrong}, in which a domain-general coreference resolution system~\cite{lee2013sieve} would link {\bf its} to {\it GSK3$\beta$} because of a generally trustworthy heuristic that the earliest named entity in the sentence is likely to be the antecedent of a pronoun if they match grammatically \cite{hobbs1978}. The domain-specific knowledge that a protein binding to itself would not be referred to this way allows us to rule this link out and instead correctly choose {\it Axin GBD}, as in \ref{ex:ds-selfbindingright}.

\begin{exe}
	\ex \begin{xlist}
		\ex[*] {\textellipsis we incubated {\it GSK3$\beta$} with excess Axin GBD protein to saturate {\bf its} \underline{binding} to GSK3$\beta$ \textellipsis\footnote{Bold face text denotes an anaphor, and italicized text denotes the antecedent chosen by the approach in discussion. Importantly, for entity resolution, we focus only on entities that participate in biochemical events; an underline denotes the anchor phrase of the corresponding event. The asterisk indicates an incorrect resolution.}}\label{ex:ds-selfbindingwrong}
		\ex[] {\textellipsis we incubated GSK3$\beta$ with excess {\it Axin GBD} protein to saturate {\bf its} \underline{binding} to GSK3$\beta$ \textellipsis}\label{ex:ds-selfbindingright}
	\end{xlist}
\end{exe}

Similarly, domain-general systems would typically make in incorrect link in example~\ref{ex:ds-studywrong}. Because such systems would not know that \underline{interaction} refers to a biochemical reaction that has molecules as its participants, ``study'' is considered as a possible antecedent (a mistake avoided by our approach in example~\ref{ex:ds-studyright}):

\begin{exe}
	\ex \begin{xlist}
		\ex[*] {{\it The only previous study} concerned the class II paired box gene Pax8, and {\bf its} \underline{interaction} with Smad3.}\label{ex:ds-studywrong}
		\ex[] {The only previous study concerned the class II paired box gene {\it Pax8}, and {\bf its} \underline{interaction} with Smad3.}\label{ex:ds-studyright}
	\end{xlist}
\end{exe}

Finally, domain general systems are able to make assumptions that do not hold in this domain. For example, in the general domain, a mention of {\bf Barack} or {\bf Obama} is likely to corefer with the more complete mention {\it President Barack Obama}. However, in the biomedical domain, entity names overlap to a great extent, and ``glycogen synthase kinase 3 beta'' is a different entity than ``glycogen''.

The contributions of this work are twofold.
First, we adapt a sieve-based coreference resolution algorithm \cite{lee2013sieve,lee2011sieve} to the biomedical domain, capitalizing on domain-specific knowledge, extending the biomedical information extraction system of \newcite{valenzuela2015odin}. Importantly, our extensions address both entity and event (i.e., interaction) coreference resolution. 
The ``sieves'' are sequentially applied heuristics for linking mentions in text, ordered from greatest to least precision and from least to greatest recall. 
Second, we show that with only seven such sieves, we achieve significant throughput gains while maintaining high precision in a large-scale DARPA evaluation based on the full content of 1000 papers.

\section{System architecture}

The sieve-based architecture described here was developed in tandem with the Open Domain Informer (Odin) system \cite{valenzuela2015odin} for event extraction. This system uses a relatively small set of human-written (and human-interpretable) rules to extract events from text. Odin, including this coreference component, is highly scalable, and can readily process thousands of papers at a rate of less than five seconds per paper, allowing the full effect of coreference resolution to be measured reliably.

Even without the coreference resolution component, Odin is capable of recognizing some relations involving coreference, because of their grammatical regularity. Specifically, it can recognize relations through relative pronouns such as {\it which}, 
as exemplified in \ref{ex:eventcoref}.

\begin{exe}
	\ex\label{ex:eventcoref} TGF$\beta$\ signaling is initiated by {\it the binding of TGF$\beta$ to TBRII}, {\bf which} 
leads to the \underline{recruitment} of TBRI.
\end{exe}

Similarly, Odin already recognizes appositive structures as in example~\ref{ex:appositive}, again due to their grammatical (appositive) structure.

\begin{exe}
	\ex\label{ex:appositive} Central to the hyperphosphorylation of Tau was the \underline{activation} of {\it GSK-3$\beta$} ({\bf glycogen synthase kinase 3 beta})\textellipsis
\end{exe}

\subsection{Assumptions}

\paragraph{Anaphors, not cataphors.} 

Although an abbreviated reference may precede a complete reference, as in example~\ref{ex:cataphor}, most anaphors point backward in biomedical texts (and in open domain texts). We make the simplifying and constraining assumption that full mentions will appear before anaphors.

\begin{exe}
	\ex\label{ex:cataphor}After {\bf its} \underline{release} from I$\kappa$B$\alpha$, {\it NF-$\kappa$B p65} can undergo post-translational modification to activate gene transcription.
\end{exe}

\paragraph{Event cardinality.}

Based on our ontology of reactions and the corresponding requirements for participants, we split $n$-ary event mentions into multiple binary events. In example~\ref{ex:cardinality}, we find two binding reactions, one each between {\it PIK3CA} and {\it Ras} and between {\it BRAF} and {\it Ras}.

\begin{exe}
	\ex\label{ex:cardinality} \textellipsis {\it PIK3CA} and {\it BRAF} are, in part, 
regulated by direct \underline{binding} to activated forms of the Ras proteins\textellipsis
\end{exe}

\paragraph{Antecedent order determines anaphor order.}

In relatively rare cases of events involving multiple anaphors, if no information hints at which anaphor corefers with which antecedent(s), we proceed left to right. This is a successful strategy in sentences like example~\ref{ex:ordering}.

\begin{exe}
	\ex\label{ex:ordering}\textellipsis while {\it over-expressed c-Cbl_1} stabilized {\it ``activated'' MLK3_2} , {\bf it_1} \underline{suppressed} {\bf its_2} capacity to promote phosphorylation \textellipsis
\end{exe}

\subsection{Sieve architecture}

Based on \newcite{lee2013sieve}, we adopt a rule-based sieve architecture for resolving coreference, in which 
deterministic processes are ordered from highest precision to lowest precision, and from lowest recall to highest recall. 
The advantages of this approach are similar to those of the rule-based architecture of Odin, and include stability and interpretability by 
humans, and high overall performance in open domain resolution. 

\begin{figure}
\includegraphics[width=\columnwidth]{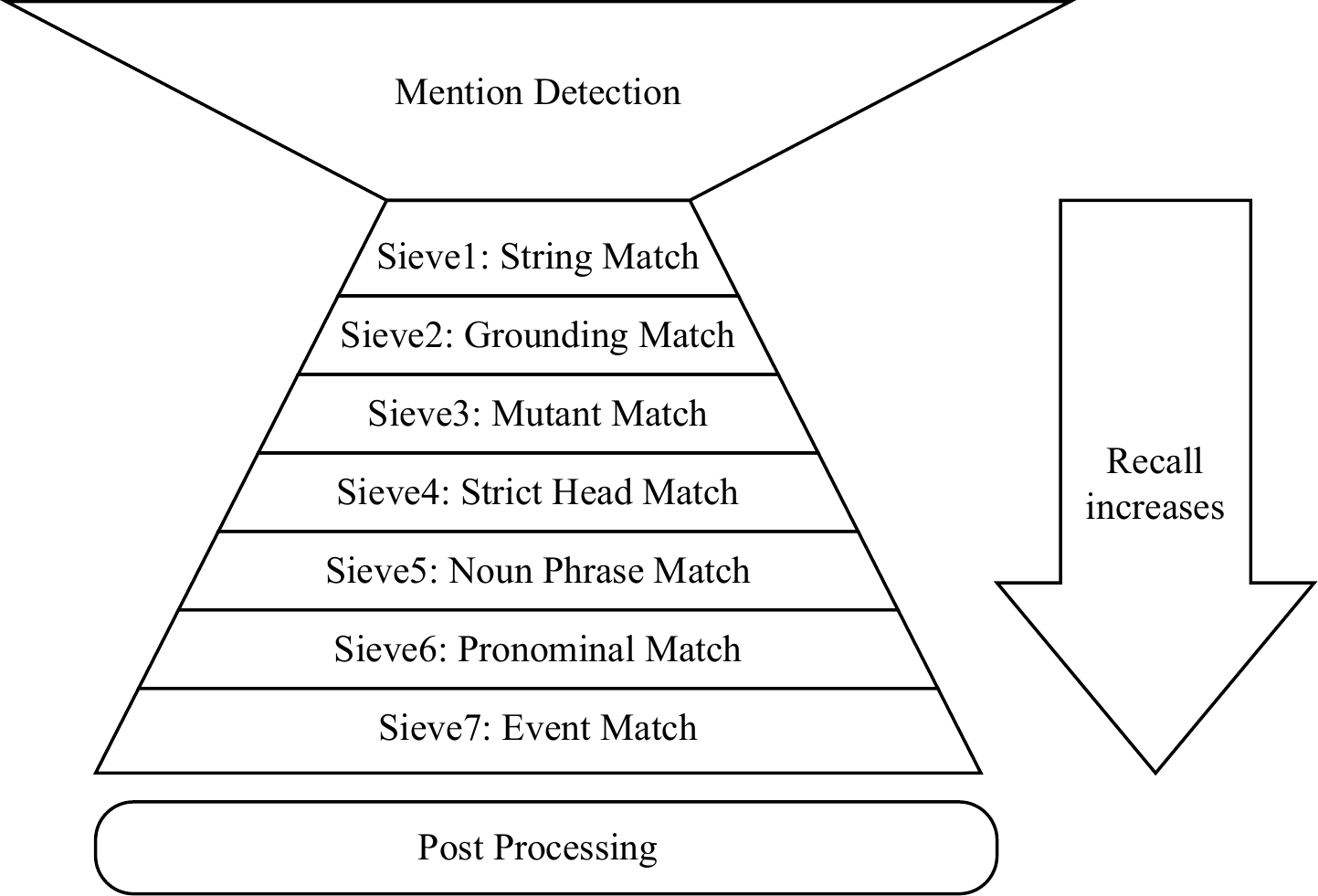}
\caption{The architecture of our biomedical coreference system.}
\label{fig:arch}
\end{figure}

Though successful in the open domain, \newcite{lee2013sieve}'s system is not well-suited to the biomedical domain, producing low-precision results. For example, it uses a sieve called {\it relaxed head match} that allows two mentions to corefer if the head of the anaphor appears anywhere in the antecedent and has no words not contained in the antecedent. This is problematic in cases such as example~\ref{ex:ikappab}, where it is insufficiently restrictive, matching {\it I$\kappa$B} with {\it I$\kappa$B kinase $\alpha$}.

\begin{exe}
	\ex\label{ex:ikappab} Two related kinases, {\bf I$\kappa$B kinase $\alpha$} (IKK$\alpha$) and IKK$\beta$, \underline{phosphorylate} {\em the I$\kappa$B proteins} \textellipsis
\end{exe} 

Here we adapt the sieves to the biomedical domain using four strategies:
\begin{enumerate}
\item We eliminate sieves that: are not applicable to the biomedical domain, are already captured by Odin rules, or are insufficiently restrictive in this domain. We eliminated the following sieves from \newcite{lee2013sieve}'s architecture: speaker identification, relaxed string match, relaxed string match, precise constructs, proper head noun match, relaxed head match. 
\item We created sieves that are specific to the biomedical domain. For clarity, we mark these sieves as ``domain specific'' in the discussion below.
\item We constrain the remaining open-domain sieves such as pronominal resolution with domain-specific constraints, e.g., forcing a pronominal anaphor to resolve to a protein when this knowledge is available. 
\item Because our ultimate goal is to re-construct protein signaling pathways, we resolve only entity mentions that participate in biochemical interactions, and incomplete event mentions. We ignore all other potential anaphors such as protein mentions that do not participate in interactions\footnote{These are still considered as candidates for entity resolution.}, and pronominal or nominal phrases in any other constructs. 
\end{enumerate}

Figure~\ref{fig:arch} shows the proposed sieve-based coreference resolution architecture for the biomedical domain. We detail all components below.

\paragraph{Mention detection.} Unlike open-domain coreference resolution, which aims to resolve all pronominal and nominal mentions in text, here we consider as anaphors only entity and event mentions that participate in fragments of protein signaling pathways. In particular, we only consider entity mentions that are arguments of relevant biochemical interactions previously extracted by Odin (e.g., phophorylations, ubiquitinations, bindings, translocations), and nominal event mentions, e.g., ``this binding''. We identify the latter using a dictionary of event trigger phrases, which must match noun phrases that also include a definite determiner. 

\paragraph{Exact string match.}

We cluster entity mentions that have the same characters in the same order. 
This is less reliable in the biomedical domain as it is elsewhere, since different proteins (in different species) may 
have the same name. However, precision remains high.

\begin{exe}
	\ex\label{ex:exactstring} \textellipsis we incubated {\it GSK3$\beta$} with excess Axin GBD protein to saturate its binding to {\bf GSK3$\beta$} \textellipsis
\end{exe}

This sieve does not boost throughput of event extraction, since the mentions it concerns are full mentions (rather than, 
say, pronouns) and thus are recognized by Odin already. However, linking full mentions as referring to the same real-world entity 
constrains later sieves and aids in assembly. For example, in example~\ref{ex:exactstring}, the link between the two mentions of GSK3$\beta$ ensures that at a later sieve, {\bf its} will not be linked to {\it GSK3$\beta$}, which would posit an incorrect chemical reaction between a protein and itself.

\paragraph{Shared grounding match (domain specific).}

The inverse problem to multiple proteins with the same name is one protein with multiple names, sometimes within a single document. We use a lookup table 
produced from large databases such as Uniprot \cite{uniprot}, containing many aliases, to cluster mentions that refer to 
the same real-world entity, whether it be a protein, a gene, a simple chemical, or cell part.
For example, in \ref{ex:appositiveagain}, {\it GSK-3$\beta$} and {\bf glycogen synthase kinase 3 beta} refer to the same entity. If this sentence were followed by ``it phosphorylates GSK-3$\beta$'', we could thereby prevent {\bf it} from coreferring to {\it glycogen synthase kinase 3 beta}.

\begin{exe}
	\ex\label{ex:appositiveagain} Central to the hyperphosphorylation of Tau was the \underline{activation} of {\it GSK-3$\beta$} ({\bf glycogen synthase kinase 3 beta})\textellipsis
\end{exe}

Similarly to the exact string match sieve, this sieve principally has the effect of constraining later sieves, in addition to aiding in assembly.

\paragraph{Mutant match (domain specific).} 

Knowing whether a protein or gene is mutated (i.e.\ altered by substituting, deleting, adding, etc., a section of the amino acids that make up a protein) is crucial to understanding reactions. The differences in reaction participation between ``wild-type'' (non-mutated) and specific mutants of a protein is often the main point of a paper. There are three types of shorthand used to refer back to fully described mutations: 
\begin{inlinelist}
\item Noun phrases such as ``the deletion mutant'' that only specify that there is a protein with a (kind of) mutation, but not which protein was mutated or which part of the protein was affected. These cases, exemplified in example~\ref{ex:mut-genericNP} are handled by our later class-based noun phrase resolution sieve.
\item Noun phrases such as ``S34A mutant'' that specify which mutation is discussed (S34A), but not which protein is being mutated. This case, exemplified in \ref{ex:mut-mutation} is similarly handled as a special case of noun phrase resolution, which is discussed later.
\item Noun phrases such as ``all six FGFR3 mutants'' that specify which protein is mutated (FGFR3), but not which mutation has taken place, as exemplified in example~\ref{ex:mut-protein}. 
\end{inlinelist} 
This last class is handled by the mutant match sieve, which tries to link an entity that has an unknown mutation to a prior mention of the same entity with the mutation spelled out.
Note that this may yield one-to-many resolution links, if the anaphor is a plural noun such as in example~\ref{ex:mut-protein}:

\begin{exe}
	\ex\label{ex:mut-genericNP} The anti-pSer34 antibody reacted with {\it AATYK1A} but not with the {\bf S34A mutant [of AATYK1A]} \textellipsis .
	\ex\label{ex:mut-mutation} \textellipsis we prepared recombinant {\em H2AX-K134A}\textellipsis The intensity of the band corresponding to histone H2AX methylation was significantly diminished in {\bf the K134A mutant} compared with that of wild-type H2AX (H2AX-WT)\textellipsis .
	\ex\label{ex:mut-protein} Cells were transfected with {\it N540K, G380R, R248C, Y373C, K650M and K650E-FGFR3 mutants} \textellipsis {\bf all six FGFR3 mutants} induced activatory ERK(T202/Y204) phosphorylation\textellipsis .
\end{exe}

\paragraph{Strict head match.}

Entities are linked if the anaphor's head word is contained in the antecedent mention and 
the anaphor mention contains only words contained in the antecedent mention (with the exception of stop words). For 
example, {\it a phosphorylated ASPP2 protein} matches {\it the ASPP2} and {\it the phosphorylated protein} but not {\it 
the activated ASPP2}. An example of a match is given in example~\ref{ex:headmatch}, where the head {\it enzyme} precedes the entity mention text.

\begin{exe}
	\ex\label{ex:headmatch} \textellipsis in {\it the enzyme guanylate cyclase}. As a result, {\bf the enzyme} \underline{becomes active} and catalyses the production of more cGMP from GTP.
\end{exe}

\paragraph{Pronominal and determiner resolution.}

Pronominal coreference is very common in biomedical literature. In fact, pronominal anaphora are the most common in the BioNLP Genia Event Extraction (GE) 2013 gold corpus, with {\it it} and {\it its} being the two most common anaphoric expressions. A typical case is shown in example~\ref{ex:foxp3}.

\begin{exe}
	\ex\label{ex:foxp3} {\it FOXP3} is an essential transcription factor \textellipsis; however, the mechanisms 
regulating {\bf its} \underline{expression} are as yet unknown.
\end{exe}

Most of the variables useful for open-domain pronoun resolution are irrelevant here, 
particularly gender, person, and animacy, since the entities and events mentioned are invariably referred to as neuter, 
3\textsuperscript{rd} person, and inanimate in English ({\it it}, {\it its}, {\it them}, etc.). However, pronoun number remains crucial, in that it denotes how many antecedents to link to.

Following \newcite{hobbs1978}, we use a simple heuristic for finding the antecedent of these expressions, starting from the beginning of the current sentence and traveling linearly rightward until an appropriate mention (or mentions) is reached. If insufficient mentions are found this way, we traverse the immediately previous sentence left to right. Unlike Hobbs, who traversed trees, we simply use word order, which is effective in most cases.

We improve this search using several domain-specific constraints. 
We exclude from this search any mentions that are participants in the current event, which prevents the system from concluding that a protein phosphorylated itself, for example. We further exclude any mentions that have been previously marked as coreferent to any mention that is a participant in the current event, as well as any mentions that we know through domain knowledge cannot be a participant in the current reaction. For example, in a phosphorylation event (the addition of a phosphate molecule to another molecule), the thing being phosphorylated must be a protein or other chemical, and cannot be a sub-cellular location such as the cytoplasm.

Furthermore, a single event mention in the text may be split into multiple events based on the cardinality of the event and of its anaphoric participants. In example~\ref{ex:copeak}, {\bf their} must refer to multiple antecedents (or to one or more plural antecedent), and each of these antecedents must participate in a binding event with FLAG-CUL4A but not with each other.

\begin{exe}
	\ex\label{ex:copeak} \textellipsis endogenous {\it BAF} and {\it emerin} consistently ``co-peaked'' in {\bf their} \underline{interaction} with FLAG-CUL4A after UV-treatment.
\end{exe}

\paragraph{Class-based noun phrase resolution (domain specific).}

Relevant entities fall into discrete classes, each with its own set of periphrastic expressions.

\begin{exe}
	\ex\label{ex:protein} \textellipsis {\it Rb} binds to E2F. {\bf The protein} also \underline{inhibits} the 
transactivation capacity of E2F.
	\ex\label{ex:rsmad} \textellipsis the receptor Smads ({\it Smad-1}, {\it Smad-5}, and {\it Smad-8}). {\bf The R-
Smads} then \underline{form complexes} with the co-Smad (Smad4) and are translocated into the nucleus\textellipsis
\end{exe}

Classes include proteins ({\it the protein}, {\it kinases}), protein families, genes, sites ({\it site}, {\it this position}), and 
simple chemicals, but periphrastic noun phrases are much more numerous, referring to subclasses and 
subsubclasses of each. 

Of the anaphora in the gold-annotated GE corpus, 41.0\% (91 of 222) are closed-class words such as {\it it}, {\it 
which}, and {\it that}. The remainder are periphrastic noun phrases that reference the 
class of the antecedent entity or event.

At each noun phrase from a constrained set such as ``the protein'', we search only for proteins, using the same search heuristic as in the pronominal resolution sieve. The matches are further constrained in that the anaphoric expressions must contain linguistic evidence that they are coreferent with some already mentioned entity, such as a definite determiner article ({\it the}) or a demonstrative word ({\it these}, {\it that}). Thus, ``a kinase'' does not match, but ``this kinase'' can.

\paragraph{Event coreference (domain specific).}

Simple event mentions such as phosphorylation events can be participants in regulations. When this occurs, they may be full mentions or they may be incomplete, as in example~\ref{ex:event}:

\begin{exe}
	\ex\label{ex:event} {\it LL-37 forms a complex together with the IGF-1R} \textellipsis and {\bf this binding} \underline{results} in IGF-1R activation \textellipsis .
\end{exe}

When an incomplete simple event mention is a participant in a regulation, we search for complete event mentions, i.e., event mentions with the expected number of arguments present, constraining our search to events of the same type. Although in principle regulation events can be participants in other regulation events, we constrain the system to one level of recursion to maintain precision, so we do not perform searches on anaphora that indicate regulation events such as {\it the promotion}.

\paragraph{Clean-up.}

The entity and event recognition process for finding anaphoric expressions casts a very wide net. Because the event rules and sieves are carefully constrained to only find appropriate anaphoric relationships, it is not harmful to recognize, for example, expletive {\it it} in expressions like {\it It is hypothesized\textellipsis}. However, it is necessary to clear any entities and events for which appropriate antecedents were not found. These are simply filtered out of the reported entities and events.

\section{Experiments}

The system was evaluated on materials from a 2015 large-scale Big Mechanism evaluation on 1000 papers. While a corpus
gold-annotated for coreference does exist for the biomedical domain, the BioNLP GE 2013 corpus \cite{kim2013ge}, we instead use the Big Mechanism evaluation corpus for two main reasons. 

First, the BioNLP corpus is not compatible with the model architectures that are central to Big Mechanism, e.g., the Biological 
Pathways Exchange Language (BioPAX) \cite{Biopax} and Biological Expression Language (BEL)\footnote{\url{http://www.openbel.org/}}.
For example, the 
BioNLP corpus makes no functional distinction between regulation and activation events, contrary to BioPAX's ontology, which 
specifies regulations (as ``controls'') but not activations. Also, BioPAX binding events require at least 2 participants, while 
BioNLP binding events allow 1 to 6 participants.

Second, the 
scale of the BioNLP corpus is necessarily limited due to its hand-curated design, with only 20 papers in its training and 
development subcorpora combined. The relative sparsity of coreference relations that contribute to event extraction 
that would otherwise be impossible makes this size insufficient. 

Because Odin does not require training, but rather is rule-based, and its rules were developed on different texts, no separation of the DARPA corpus into training and testing was necessary.

\subsection{Evaluation}
\label{sec:exp}

Given that this 1000-paper corpus is not annotated, a recall measure is not possible, as there is no gold set of events to be 
extracted. Following the DARPA Big Mechanism evaluation, then, we measure throughput, defined as the number of events (or interactions) mentions extracted, as an approximation of recall.\footnote{In a slight departure from DARPA's definition of throughput, here we count regulations and biochemical reactions as distinct event mentions. In the DARPA evaluation, regulations were collapsed with the corresponding controlled reaction into a single event mention.}

Likewise, precision must be hand-coded on a random sample of the program's output. For the overall system, this was performed 
by independent raters chosen by DARPA, using a metric called {\it generous precision}. In generous precision, the accuracy of a 
single event extraction is 1 if it is considered useful, i.e., if it is the correct event type and has at least one correct participant, and 0 
otherwise. The overall generous precision is the average accuracy over evaluated events. Because too few of the events so evaluated 
resulted from coreference resolution to sufficiently evaluate the coreference system itself, we performed a similar evaluation on 
100 randomly selected events extracted using coreference resolution, using the same metric.

\subsection{Results}

\begin{table}
\centering
\begin{tabular}{l l}
\toprule
{\it Throughput mentions}	\\
\hspace{1em}Odin			& 46,234	\\
\hspace{1em}Coreference alone			& 1,492	\\
\hspace{1em}Odin + coreference		& 47,726	\\
\midrule
{\it Generous Precision}	\\
\hspace{1em}Odin + coreference			& 74.2\%	\\
\hspace{1em}Coreference alone			& 68.0\%	\\
\bottomrule
\end{tabular}
\caption{Throughput and precision for the Odin system with and without the coreference resolution component.}
\label{tab:results}
\end{table}

The results summarized in Table \ref{tab:results} show that our sieve-based coreference system contributed a 3.22\% increase in 
throughput. A separate analysis of the BioNLP 2013 GE gold-annotated corpus shows that the maximum possible contribution of 
coreference is 8.9\% (under slightly different definitions of reaction types). 
The BioNLP figure includes phenomena commonly included under the umbrella of ``coreference'', namely the anaphor {\it which} followed by a relative clause and the appositive structures described earlier, which the original Odin system already addressed.

\begin{table}
\centering
\begin{tabular}{l l}
\toprule
{\it Error source}	& {\it Portion of errors}	\\
\midrule
Named entity recognition	& 14\%	\\
Event recognition	& 36\%	\\
Coreference resolution	& 50\%	\\
\bottomrule
\end{tabular}
\caption{Error analysis by source of the error. This analysis was performed over all precision errors produced by the coreference resolution component.}
\label{tab:errors}
\end{table}

Importantly, the precision of the coreference system approaches that of the Odin system generally (68\% vs. 74\%). This is encouraging: the 74\% generous precision was the second highest precision score reported in this evaluation (the highest was 75.8\%, but at a lower throughput). This demonstrates that, despite its simplicity, the proposed coreference resolution approach has state-of-the-art performance.

\begin{table}
\centering
\begin{tabular}{ @{} p{0.25\columnwidth} p{0.7\columnwidth} @{}}
\toprule
{\it Error type}	& {\it Example}	\\
\midrule
Named entity recognition	& Moreover, although {\it RoR-siRNA} alone was able to increase the p53 level, {\bf it} did not \underline{cause} p53 phosphorylation or acetylation\textellipsis	\\
Event recognition	& {\it GST and GST-hBex2} fusion proteins were used to test {\bf their} \underline{interaction} with EGFP-tagged LMO2	.\\
Coreference resolution	& \textellipsis one of two adaptor proteins that varied in {\bf their} reported \underline{interactions}: (1) an adaptor protein that was not known to directly associate with the neurotrophin receptor or the channel ({\it Grb10}), and (2) an adaptor protein that was known to bind to Y\textsuperscript{490} (NPQpY motif) of the neurotrophin receptor ({\it nShc}).	\\
\bottomrule
\end{tabular}
\caption{Examples of errors in coreference resolution by class of error. The correct antecedent is italicized for clarity, but the antecedents were not recognized.}\label{tab:error-examples}
\end{table}

A further error analysis summarized in Table~\ref{tab:errors} shows that only half of the precision errors are due to actual faulty coreference resolution, while the remaining half are due to errors of 
the general system, e.g., failure to recognize named entities for antecedents, failures of the 
event detection rules, and incorrect syntactic analysis on complex sentences.

Table~\ref{tab:error-examples} shows an example of each of the types of error found. The first example fails because {\it RoR-siRNA} was not recognized as an entity; replacing it with an entity known to the event extraction system causes the correct antecedent to be recovered. In the second case, the correct result is that {\it GST} and {\it GST-hBex2} should each interact with {\it LMO2}. Because the event extraction system did not recognize that LMO2 was involved in the event, although it recognized the entity LMO2, the coreference system was left to resolve only {\bf their} \underline{interaction} and produced a single event with two participants, {\it GST} and {\it GST-hBex2}. Finally, there is a case in which the assumptions of the coreference system have been violated: the fully specified mentions {\it follow} rather than {\it precede} the anaphor. This situation is difficult to detect, but future work may extend to recognizing list structures such as that in this example, which may indicate cataphora (in which the full mention follows the anaphoric expression) in a reliable fashion.

\subsection{Mutation evaluation}

Mutation resolution was a late addition to the coreference system and so evaluated separately for precision. Three raters examined a total of 77 entity and event extractions that included at least one mutant that required coreference resolution. For these entity and event mentions, a point was awarded for each correct resolution, a half point if the named antecedent was the correct protein but had the wrong modifications (e.g., it was not mutated), and zero points if the named antecedent was incorrect. This is similar to the ``generous precision'' above in that it gives credit for less than maximally informative resolutions. The precision under this definition was 75.7\%. Of the errorful resolutions, 79\% were due to failures of named entity recognition in sentences like example~\ref{ex:longmut} in which the mutation description was complicated or used nonstandard notation. 

\begin{exe}
	\ex\label{ex:longmut} When {\it RUFY1 was further truncated from the C-terminus [RUFY1(1-420)]}, the {\bf truncation mutant} could not \underline{bind} to either Rab14 or Rab4.
\end{exe}

The remaining 21\% of errors were due to coreference resolution failures, but 67\% of these were because a full antecedent was simply never mentioned in the prior text. In example~\ref{ex:no-antecedent}, a sentence from an abstract mentions {\bf the double mutant} which has not yet been defined. In a later section of the text, the mutant is fully described as Nudel\textsuperscript{N20/C36}. This violates the system's general assumption that a full antecedent will precede the anaphor, and by several paragraphs.

\begin{exe}
	\ex\label{ex:no-antecedent} In contrast, the wild-type Nudel and {\bf the double mutant} that \underline{binds} to neither protein are much less effective.
\end{exe}

All in all, this analysis indicates that the resolution of protein mutants performs comparably (slightly higher in fact) to the rest of the biomedical coreference resolution system. To our knowledge, this is the first evaluation of a system linking mentions in text that are more or less specific references to mutant proteins. Furthermore, similar to the analysis in the previous sub-section, this error analysis supports the observation that only a small percentage of the errors are inherent limitations of the proposed coreference resolution architecture.

\section{Discussion}

\begin{figure*}[!htbp]
\center
\includegraphics[width=0.6\textwidth]{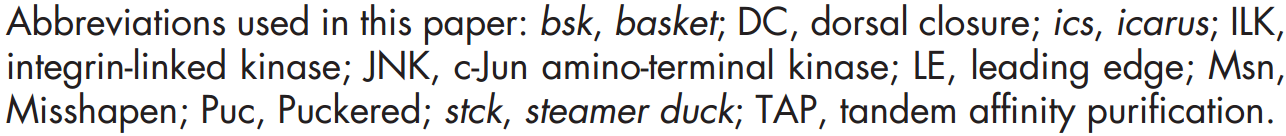}
\caption{A note containing explicit abbreviations from \newcite{kadrmas:2004}.}\label{fig:aliases}
\end{figure*}

This system, while productive, does not yet take maximal advantage of the information in the text. Future work and ongoing improvements are discussed next.

\subsection{Alias resolution}

Terminology is quite inconsistent between papers, because authors  often introduce name aliases for proteins or mutations such as ``E3 ligase BRAP (also referred to as IMP)''. These aliases will hold for the duration of the paper, but they generally do not appear in knowledge bases such as Uniprot \cite{uniprot}, which complicates resolution. In this case, the nonce equivalence of two names is informative for building models of interactions, in which the precise entity referred to must be known. In future work, we will implement patterns for the detection of these aliases when they are introduced, e.g., ``also referred to as''. Additionally, some papers contain a special section or note naming these aliases or abbreviations, as shown in Figure~\ref{fig:aliases}. Our coreference resolution system is currently being expanded to recognize text that marks two names equivalent in this way. 


\subsection{Informative ellipsis}

Because of the mechanistic nature of the events in this domain, authors may omit a great deal of information. For 
example, ellipsis, once reconstructed, can easily double the number of reactions detectable in a clause.

\begin{exe}
	\ex\label{ex:ellipsiswomod} The dephosphorylated {\it Axin} \underline{binds} $\beta$-catenin less efficiently than 
{\bf the phosphorylated form [of Axin} binds $\beta$-catenin].\footnote{Material between square brackets is added to 
complete the ellipsis and is not in the original text.}
	\ex	\begin{xlist}
		\ex[] {JNK-I revealed a stronger \underline{inhibitory} effect on IL-6 expression than the PKC-I [revealed an inhibitory effect on IL-6 expression].}\label{ex:verbalellipsis}
		\ex[*] {JNK-I revealed a stronger \underline{inhibitory} effect on IL-6 expression 
than [JNK-I revealed an inhibitory effect on] the PKC-I.}\label{ex:verbalellipsiswrong}
		\end{xlist}
\end{exe}

In example \ref{ex:ellipsiswomod}, proper understanding of the elided material allows us to capture the binding of 
phosphorylated Axin to beta-catenin as well as the negative regulation of dephosphorylation of Axin on the binding of 
Axin and $\beta$-catenin. To accomplish this, it is necessary to match the protein antecedent (Axin) but not the 
modification (dephosphorylation).

Similarly, the regulation of PKC-I on the IL-6 is captured only using ellipsis in example \ref{ex:verbalellipsis}. We must 
discount competing interpretations of the ellipsis such as that in example \ref{ex:verbalellipsiswrong}, which is a 
grammatically acceptable but factually incorrect alternative.

\subsection{Reference to latent entities}

The outputs of biochemical reactions are redundant with the reaction type and participants, so they are also often 
elided. For example, a (protein) binding reaction results in a (protein) complex made up of the participants in the 
binding. The complex may then be referred to without being explicitly named. This is essentially an anaphor without an 
explicit antecedent---a latent entity.

\begin{exe}
	\ex\label{ex:latent1} TGF$\beta$ signaling is initiated by the binding of TGF$\beta$ to TBRII. {\bf The [resulting] 
complex} [{\it TGF$\beta$:TBRII}] then \underline{recruits} TBRI.
\end{exe}

In example \ref{ex:latent1}, {\bf The complex} refers to {\it TGF$\beta$:TBRII}, the outcome of the binding described in the 
previous sentence. It is this complex that recruits (i.e., binds with) TBRI to form the complex TGF$\beta$:TBRII:TBRI.

This extension requires a model of the output of each event that has been found, work which is currently underway.

\section{Conclusion}

This work provides one of the first empirical measurements of the impact of entity and event coreference on large-scale event extraction in the biomedical domain. Furthermore, while prior work by \newcite{miwa2012coref} discusses the ideas of coreference in the biomedical domain, here we offer a concrete algorithm which leverages specific domain constraints.

This work was motivated by the observation that open-domain coreference resolution systems perform poorly in the biomedical domain because open-domain algorithms are insufficiently restrictive in this domain. To address this, we modified the sieve-based approach of \newcite{lee2013sieve} to infuse domain knowledge by: (i) removing open-domain sieves that do not transfer well to the biomedical domain; (ii) adding novel, domain-specific sieves, and (iii) constraining the remaining open-domain sieves with domain-specific restrictions that control which anaphors to resolve and which candidates to consider during the resolution process. 

We offer quantified results for a state-of-the-art event extraction system extended with the resulting coreference resolution approach, which show increased throughput at comparable precision when coreference resolution is used.

\section{Software}
The software described here is open-source and available at \url{http://github.com/clulab/reach}.

\section*{Acknowledgments}
This work was funded by the Defense Advanced Research Projects Agency (DARPA) Big Mechanism program under ARO contract W911NF-14-1-0395. We thank MITRE for implementing the evaluation discussed in Section~\ref{sec:exp}
\section{References}

\bibliographystyle{lrec2016}
\bibliography{coref}

\end{document}